\documentclass[letterpaper, 10 pt, conference]{ieeeconf}  

\IEEEoverridecommandlockouts                              

\usepackage{ifpdf}
\usepackage{amsmath,amsfonts}
\usepackage{amssymb}  
\usepackage{cases}
\usepackage{algorithmic}
\usepackage{algorithm}
\usepackage{array}
\usepackage[caption=false,font=footnotesize]{subfig}
\usepackage{textcomp}
\usepackage{url}
\usepackage{verbatim}
\usepackage{graphicx}
\usepackage{cite}
\usepackage{booktabs}
\usepackage[export]{adjustbox}
\usepackage{bm}
\usepackage{nccmath}
\usepackage{mathtools}
\usepackage{hyperref}
\hypersetup{hidelinks}
\usepackage{tensor}
\usepackage[dvipsnames]{xcolor}
\usepackage{units}
\usepackage{tikz-cd}
\usepackage{bbm}
\usepackage{tikz}
\usepackage{multirow}
\usepackage{accents}
\usepackage{soul}
\usepackage{optidef}
\usepackage{threeparttable}
\usepackage{layouts}

\DeclareMathOperator{\trace}{trace}

\DeclareMathOperator{\diag}{diag}

\DeclareMathOperator{\convhull}{convhull}

\newcommand{\bmm}[1]{\bm{\mathrm{#1}}}
\newcommand{\tsum}[0]{\textstyle\sum}
\newcommand*{\myDots}{\ifmmode\mathellipsis\else.\kern-0.13em.\kern-0.13em.\fi} 
\newcommand{\tj}[0]{\mathrm{j}}
\newcommand{\tm}[0]{\mathrm{m}}

\newcolumntype{R}[1]{>{\hspace{-15pt}\raggedleft\let\newline\\\arraybackslash\hspace{0pt}}m{#1}}

\title{\LARGE \bf
Control- \& Task-Aware Optimal Design of Actuation System   \\ for Legged Robots using Binary Integer Linear Programming
}

\author{Youngwoo Sim, Guillermo Colin and Joao Ramos 
\thanks{This work is supported by the National Science Foundation via grant CMMI-2043339.}
\thanks{The authors are with the Department of Mechanical Science and Engineering at the University of Illinois at Urbana - Champaign, USA.{\tt\small (sim17@illinois.edu)}}%
}

\begin{document}

\maketitle
\pagestyle{empty}


\begin{abstract}
Athletic robots demand a whole-body actuation system design that utilizes motors up to the boundaries of their performance. 
However, creating such robots poses challenges of integrating design principles and reasoning of practical design choices. This paper presents a design framework that guides designers to find optimal design choices to create an actuation system that can rapidly generate torques and velocities required to achieve a given set of tasks, by minimizing inertia and leveraging cooperation between actuators. The framework serves as an interactive tool for designers who are in charge of providing design rules and candidate components such as motors, reduction mechanism, and coupling mechanisms between actuators and joints. A binary integer linear optimization explores design combinations to find optimal components that can achieve a set of tasks. The framework is demonstrated with 200 optimal design studies of a biped with 5-degree-of-freedom (DoF) legs, focusing on the effect of achieving multiple tasks (walking, lifting), constraining the mass budget of all motors in the system and the use of coupling mechanisms. The result provides a comprehensive view of how design choices and rules affect reflected inertia, copper loss of motors, and force capability of optimal actuation systems. 
\end{abstract}

\section{Introduction}

Athletic robots such as the MIT cheetah and Cassie have emerged to show superior performance to their semi-static counterparts. These machines possess the ability to execute agile movements, such as utilizing momentum to jump or push heavy objects. Achieving such athletic motions requires a comprehensive approach to whole-body control  and systematic actuation system design \cite{MIT_humanoid_backflip, cheetahLanding, artemis_thesis}, wherein multiple actuators are built to operate cooperatively across the robot's body. While the literature has made significant strides in whole-body control of humanoids, the design problem pertaining to high-DoF systems still demands further investigation. 

The ability of athletic robots to generate large forces and acceleration is harnessed by incorporating design principles at two distinct layers of the actuation system. At the actuator level, the quasi-direct drive (QDD) paradigm \cite{ImpactMitigation} suggests minimizing the inertia of the limbs. Using smaller gear ratios minimizes the reflected inertia of individual actuators and reduces inertial torques required to swing the limbs. In cases where gear ratios are not large enough to amplify torque sufficiently, this can be addressed at the system level by employing cooperative (coupled or parallel) actuation \cite{sim2022tello}, where actuators are coupled together to combine their torque output and meet the torque requirements of specific joints.

However, as the number of actuators of a system grows, designers face multi-faceted issues in integrating the aforementioned principles. First, the design considerations for multi-actuator systems, such as humanoid legs, are not trivial. Design principles often
provide general intuition but leave specific design choices to the discretion of designers. Consequently, designers need to analyze the trade-offs between component choices
and their impact on performance metrics in a multi-dimensional
sense. A popular choice of performance metrics are, namely, inertia matrix which measures kinetic energy of the entire system \cite{khatib1995inertial} and force capability which measures available force that end-effectors can exert on the environment \cite{chiacchio1997force}. In legged robot design, coupled actuation has shown potential benefits of amplifying joint torques without increasing reflected inertia extensively \cite{sim2022tello}. However, under coupling, modifying even a single gear ratio out of
many actuators affects torque and velocity capabilities of other actuators, which complicates design analysis. Second, it is difficult to smoothly parameterize characteristics
of available design candidates. Designers typically source motors or reduction mechanisms from the market, where there can be significant jumps in properties between available options. This demands designers reflect the discontinuities in design studies, even complicating the design process.

\begin{figure}[t]
\centering
\includegraphics[width=\linewidth]{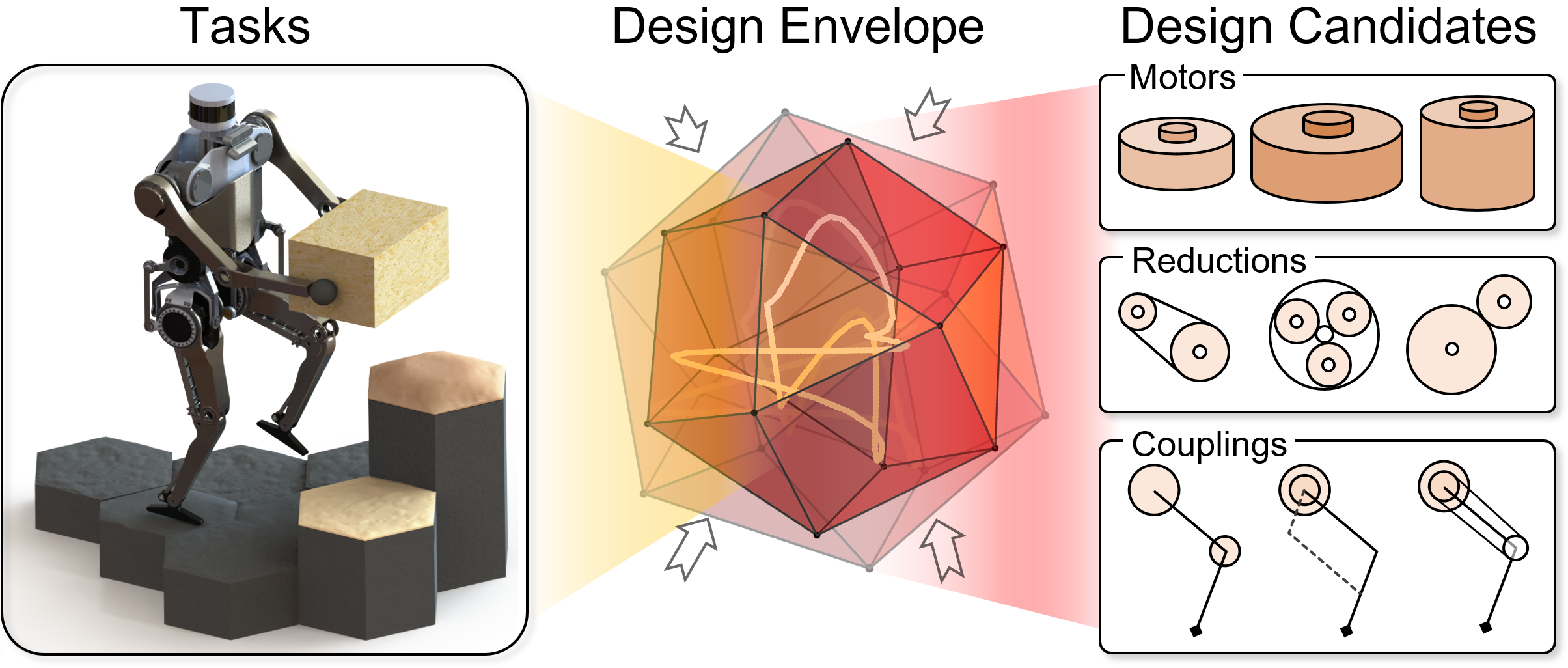}
\caption{Conceptual overview of the proposed design optimization of multi-actuator system. Video: \url{https://youtu.be/DEXYp0TsqRI}}
\label{fig:front}
\end{figure}

Fortunately, a broad spectrum of methods exists to address the challenges associated with designing multi-DoF systems. One straightforward approach involves defining high-level tasks and identifying optimal component choices, such as gear ratios, motors, and limb lengths, with the goal of optimizing system performance metrics such as power consumption \cite{kaistMINLP, chadwick2020vitruvio, pawlus2015drivetrain} or force capability \cite{gripperCoOpt}. Another family of methods is called the co-design paradigm, wherein the design and control aspects are jointly optimized through the addition of a controller and simulation of design cases \cite{ControlCoDesign}. This co-design approach emphasizes control-related aspects such as sensitivity \cite{ha2017joint} and robustness \cite{LargeScaleADMM, digumarti2014concurrent}. 

However, existing optimal design tools lack certain crucial aspects of the design process, often yielding theoretical designs that lack practicality in real-world applications. First, the expert knowledge of designers, which includes non-parametric design rules, proves essential in humanoid design. For example, the actuation system design in humanoids involves intricate space optimization to accommodate multiple actuators within the robot's body. These considerations are better defined as logical constraints rather than parametric ones. Secondly, designers typically rely on optimal design software to gain a comprehensive view of multiple design scenarios, rather than seeking a singular theoretically optimal solution. Moreover, designers engage in an iterative process of obtaining optimal designs from software and making necessary modifications. In this process, design tools assist in reasoning and justifying these modifications. Hence, the software employed must be relatively fast and capable of incorporating human intuitions. Regrettably, current state-of-the-art frameworks heavily rely on computationally intensive algorithms, such as genetic algorithms, or the number of optimization variables is limited. 

This paper proposes a \textit{design tool} that assists designers in creating high-DoF actuation systems, addressing the aforementioned issues. First, we want a design tool that guarantees a system design that is capable of various humanoid tasks such as carrying objects, weight-lifting, and/or running. 
However, it is a chicken-and-egg problem; without a robot design, one cannot reason about tasks and vice versa. Thus, tasks or motions are planned based on a reduced-order model (ROM) of a robot consisting of high-level design parameters such as overall mass, inertia, and center of mass (CoM) height with fixed morphology and ordering of joints. These assumptions are consistent with existing motion planners based on single rigid body dynamics (SRBD) \cite{batke2022optimizing} or centroidal dynamics \cite{MIT_humanoid_backflip}, allowing us to use similar ROM-based motion planners. 
Second, in this framework, designers are encouraged to use their expert knowledge to populate and curate libraries of design candidates such as motors, gear ratios, and coupling mechanisms between actuators and joints. Moreover, the high-level intuition of designers is translated into design rules, which designers can experiment with to observe how optimal solutions behave under these rules. Third, our objective is to create dynamic humanoids capable of rapid limb acceleration and impact mitigation. To achieve these goals, \cite{ImpactMitigation} suggests that minimizing the reflected inertia of actuation systems is the key, which sets the objective of the proposed design optimization to choosing components from design libraries that minimize reflected inertia while ensuring task completion. Lastly, to solve the optimization more efficiently, the optimal design problem is translated into a binary integer linear program which yields optimal solutions faster than existing methods.  

This framework is demonstrated by design studies of a biped with 5-DoF legs focusing on the effects of 1) achieving a set of tasks, 2) employing actuator-joint couplings (parallel mechanism) and 3) limiting the mass budget of all motors in a system. The optimization study was given two tasks that involve different joint usage patterns: walking and weight-lifting in a snatch style. Next, design libraries of motors, gear ratio, and coupling mechanisms were compiled from existing humanoids and unprecedented-but-practical designs for comparison. 
From the results of 200 optimal design studies with varied motor mass budgets and usage of couplings, several trends in reflected inertia, force/torque capability, and copper loss of motors were observed.

The contribution of the proposed framework is as follows. First, this tool allows designers to quickly navigate vast design spaces and reveals a comprehensive overview of optimal multi-actuator system designs for legged robots. Second, 
the framework serves as a reasoning tool for high-DoF system design. Designers can quickly run a large number of optimal studies to observe how optimal solutions respond to modifying tasks, design libraries, or design rules. For example, providing a larger set of tasks would result in general-purpose humanoid design whereas providing a single task would yield a more dedicated system design. Moreover, novice engineers can update components available in the market to create better designs, while experts can explore combinations of novel mechanisms for comparison.

\section{Background: Representation of Design Library \& Combinations} \label{sec:background}

The binary variables are efficient tools for handling two essential aspects of design: 1) determining whether to select or exclude component candidates from libraries and 2) expressing combinations of components. 

\textbf{Definition 1}: \textit{(Binary Combination)} Let $\bmm{p}\!=\!\{p_1, \cdots, p_n\}$ be a library of $n$ objects, $\bmm{x}\!=\!(x_1, \cdots, x_n)\!\in\!\mathbb\{0,1\}^n$ be a vector of binary variables. An arbitrary singular choice of an object $p(\bmm{x})_B$ out of the library $\bmm{p}$ is obtained by a \textit{binary combination of} $\bmm{p}$ \textit{and} $\bmm{x}$ as follows,
\begin{align}
&p (\bmm{x})_B \coloneqq \tsum_{i=1}^n p_i x_i  \:\:\:
\\ \Longrightarrow \:\:\: &\tsum_{i=1}^n x_i = 1 \quad (\textrm{exclusivity}).   
\end{align}
For instance, a selection of $p_j$ is achieved by assigning 1 to the corresponding variable 
$(x_j\!=\!1)$ and 0 to the rest $(x_k\!=\!0, k\!\neq\! j, k\!\in\![n])$. The definition itself implies an exclusivity constraint that enforces a single choice. 
It should be noted that the objects $p_i$ can be scalars such as maximum torque or matrices like the Jacobian of coupling mechanism that maps actuator velocities to joint velocities.

\textbf{Remark 1}: \textit{(Linearization of Multilinear Polynomials)} Products of binary variables commonly arise when combining components; e.g., an actuator is a combination of a motor and a reduction mechanism. A useful property of the product of binary variables is that it can be perfectly linearized (as opposed to linearly approximated). The product of binary variables $x$ and $y$, denoted as $xy$, is linearized by introducing a new binary variable $z$ in place of $xy$ and adding the following linear constraints:
\begin{equation}
z\le x, \quad
z\le y, \quad
z \ge x + y -1.
\end{equation}
For linearization of general multilinear forms, refer to \cite{binaryLinearization}. 
\section{Optimal Design of Transmission System for Humanoid Robots}

\subsection{Overview}
We introduce a design optimization tool that finds mechanical components for the actuation system of athletic humanoids. Most importantly, this framework is composed as an interactive tool for designers and addresses difficulties in designing a high-DoF actuation system. Hence, the framework expects designers to provide their knowledge and intuition and run several iterations to gain an understanding of optimal designs of humanoids. 

\begin{figure}[t]
\centering
\includegraphics[width=0.98\linewidth, clip, trim=0mm 102mm 254mm 0mm]{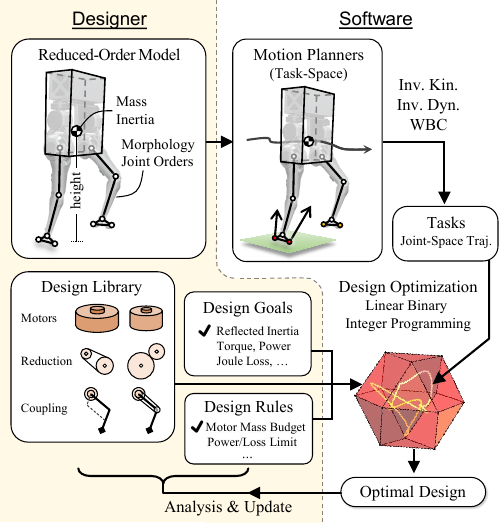}
\caption{Overview of design optimization of high-DoF actuation system for humanoid robots}
\label{fig:workflow}
\end{figure}

The workflow of the proposed tool is divided into three steps. (Fig.\,\ref{fig:workflow}). The first step is to prepare key ingredients for design optimization; 1) ROMs for the generation of task trajectories, 2) a set of task trajectories generated by motion planning software using ROMs, 3) libraries of candidate components to create the robot, 4) relevant design rules, and 5) design goals.
As to the preparation of task trajectories, first, ROMs are created from known kinematics and parameters of high-level design envelopes and behavior targets set by designers (e.g., overall mass, inertia, CoM height or desired velocity) depending on the tasks. Then, using ROM-based motion planning approaches (e.g., SRBD-based trajectory optimization \cite{ahn2021versatile} or kino-dynamic planning \cite{chignoli2021humanoid}), task-space trajectories of desired tasks are obtained. Finally, the task-space trajectories are converted to joint-space trajectories using inverse kinematics, inverse dynamics, or whole-body controllers. 
It is important to note that the ROM assumes fixed limb lengths and joint ordering. This is because humanoids are typically designed to have similar proportions and kinematics to humans. Also, as there exist only a few practical joint orderings that mimic human limb kinematics, design studies for every joint ordering can be carried out. It is also assumed that the actuators are placed very closely to the center of the robot (proximal actuation) and actuator dynamics is neglected. Hence, the actuator placements and their inertial torques do not affect the dynamics of the robot. Lastly,
the framework does not take into account the structural design of limbs because we assume that the apparent inertia and mass could be significantly minimized compared to the reflected inertia of the actuation system.   

The second step of the framework is to integrate tasks, design library, and design rules and translate them to linear binary integer programming. The overarching concept is expressed as follows with $\bmm{x}$ as design choices of mechanical components, 
\begin{mini!}|s|  
{\bmm{x}}{\textrm{ReflectedInertia}(\bmm{x}), \label{eqn:opti:obj}}    {\label{eqn:opti}}{} 
\addConstraint{\bmm{x}\in\textrm{DesignLibrary} \label{eqn:opti:lib}}
\addConstraint{\textrm{TaskTrajectory}\subseteq \textrm{OperationDomain}(\bmm{x}) \label{eqn:opti:task}}
\addConstraint{\textrm{DesignRules}(\bmm{x}) \label{eqn:opti:rule}.}
\end{mini!} 
The contents of \eqref{eqn:opti} are briefed as follows. 

\textbf{Control-Awareness\,\eqref{eqn:opti:obj}}: 
The objective is to minimize the reflected inertia of the entire actuation system perceived at the joint-level (joint-space reflected inertia). This objective aims to improve force control performance, enhance the system's ability to regulate external impacts and enable agile limb movements \cite{ImpactMitigation}. The objective aligns with the QDD paradigm, although designers have the flexibility to modify it to prioritize energy efficiency or minimum torque expenditure, depending on their interest.

\textbf{Discrete Design Space\,\eqref{eqn:opti:lib}}:
In this framework, designers are responsible for providing and modifying libraries of candidate motors, reduction mechanisms, and actuator-joint couplings. An optimization solver then searches for the optimal design by exploring all possible combinations of provided components. Designers are encouraged to use actuator-joint coupling mechanisms from existing robots, utilize their expert knowledge to curate practical component candidates, or include novel components whose utility is in question.

\textbf{Task-Awareness\,\eqref{eqn:opti:task}}:
Robots are expected to achieve multiple tasks. This framework ensures the feasibility of these tasks within the capability of motors. 
In other words, the given tasks are mapped to trajectories of motor torques and velocities. Then, these trajectories are constrained to remain within the motor limits such as voltage and thermal limit (operation domain) as in Fig.\,\ref{fig:inertiaEllipsoidCapaPolytope}(b).
The safety of a controller can be tuned by parameters that define allowable closeness between extremes of task trajectory and the boundary of the operating domain. 

\textbf{Design Rules\,\eqref{eqn:opti:rule}}:
Additional constraints can be incorporated into the optimization process, such as limiting the sum of motor mass for a subset of joints or limiting the copper loss on motors for a certain duration. As long as the constraints are multilinear polynomial, they can be perfectly linearized (see Sec.\,\ref{sec:background}). Although not presented in this paper, logical constraints can be imposed for compact arrangement of large components such as motors.

As the last step of the workflow, designers need to analyze the resulting optimal solutions and modify the aforementioned ingredients for design optimization. For instance, if the optimization is infeasible due to high torque requirements, one can expand the design library by adding more torque-dense motors, relax the design rules, or relax the task requirements (e.g., smaller total mass, smaller desired CoM velocity of the ROM).

\subsection{Maximizing Force Control Bandwidth}

Smaller reflected inertia of actuators contributes to faster limb swing or acceleration and improves the ability to mitigate impacts from contact by quickly retracting from contacts (e.g., ground contact). Both capabilities are critical requirements for dynamic humanoids. 

Hereafter, the goal of minimizing reflected inertia of the actuation system is reconnected with the earlier optimization sketch \eqref{eqn:opti} to formulate objective functions using binary variables. Although \cite{ImpactMitigation} suggests minimizing reflected inertia in task-space, obtaining such expression using binary variables in linearizable form is prohibitive due to rational forms arising from matrix inversions \cite[Eqn.\,(17)]{khatib1995inertial}. As a proxy to task-space inertia, a joint-space reflected inertia matrix is used. The downside of using this proxy is that it neglects the lever effect of limb lengths to joints which is beyond the scope of this paper. Nevertheless, some relevant observations are stated in Sec.\,\ref{sec:task_joint}.

Let us build the joint-space reflected inertia matrix of an actuation system consisting of $n_d$ joints (or actuators) from libraries of motors, gearboxes, and actuator-joint couplings with associated binary variables $\bmm{x}_i\!\in\!\mathbb{B}^{n_{{m}_i}}$, $\bmm{y}_i\!\in\!\mathbb{B}^{n_{g_{i}}}$, $i\!\in\![n_d]$, and $\bmm{z}\!\in\!\mathbb{B}^{n_c}$, where $n_{(\cdot)}$ is the number of candidates in the respective libraries. The reflected inertia of an $i$'th actuator is the product of binary combinations of rotor inertia $I_i^r(\bmm{x}_i)_B$ and gear ratio squared $N^2_i(\bmm{y}_i)_B$, 
\begin{equation}
    I_i^a(\bmm{x}_i,\bmm{y}_i) = I^r_i (\bmm{x}_i)_B N_i^2 (\bmm{y}_i)_B.
\end{equation}
Next, the reflected inertia of each actuator forms a diagonal matrix of actuator-space reflected inertia $\bmm{H}_{\textrm{r}}^a (\bmm{x},\bmm{y}) = \diag({I}^a_1, \dots, {I}^a_{n_d})$. This inertia matrix is 
projected on to joint-space by a binary combination of actuator-joint coupling Jacobian $\bmm{C}(\bmm{z})_B$ which represents a mechanical connection between actuators and joints. The coupling Jacobian maps actuator velocities $\dot{\bmm{\psi}}\in\mathbb{R}^{n_a}$ to joint velocities $\dot{\bmm{q}}\in\mathbb{R}^{n_a}$ as $\bmm{C}:\dot{\bmm{\psi}} \rightarrow \dot{\bmm{q}}$. 
The joint-space reflected inertia is as follows,
\begin{gather}
    \overline{\bmm{H}}^{\tj}_r (\bmm{x},\bmm y, \bmm z, k)= \bmm{C}^{-\top}\bmm{H}_{\textrm{r}}^a \bmm{C}^{-1},
\end{gather}
where $k$ denotes the time instance at which the inertia is evaluated along the given tasks and $\bmm x$, $\bmm y$ are lifted binary variables of $\bmm{x}_i$, $\bmm{y}_i$.

\begin{figure}[t]
\centering
\includegraphics[width=0.9\linewidth, clip, trim = 0mm 142mm 230mm 0mm]{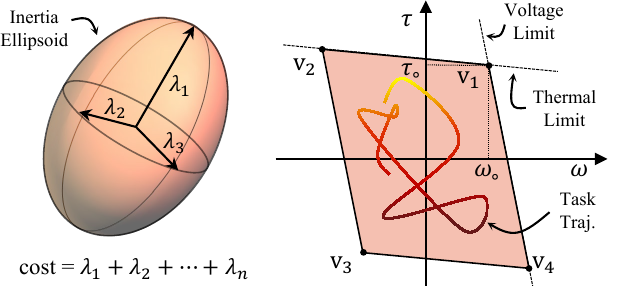}
\caption{(a) The size of reflected inertia matrix is measured as the sum of semi-axes lengths which is obtained by trace operation (b) Task trajectory projected on to motor-level is constrained to remain within the operation domain of a motor defined by vertices $\bmm{v}_i$.}
\label{fig:inertiaEllipsoidCapaPolytope}
\end{figure}

Lastly, the \textit{size of the inertia matrix} is measured using the $\trace()$ operator. The trace of an inertia matrix is the sum of its eigenvalues which relate to the lengths of semi-axes of the inertia ellipsoid (Fig.\,\ref{fig:inertiaEllipsoidCapaPolytope}(a)). Hence, the trace operator allows the optimizer to minimize reflected inertia on every dimension until motor mass becomes too small and, subsequently, a motor does not have sufficient torque to achieve given tasks.

\subsection{Task Feasibility}

Task feasibility assures that the desired task trajectories are accommodated within the capabilities of the motors. The torque and velocity capability or an operation domain of a motor is approximated as a convex polygon $\convhull(\bmm{v}_1,\dots,\bmm{v}_n)$ (colored polygon in Fig.\,\ref{fig:inertiaEllipsoidCapaPolytope}(b)). This polygon comprises vertices $\bmm{v}_i = (\alpha_i \omega_{\circ}, \beta_i \tau_{\circ})$, whose coordinates are defined by motor parameters $\omega_{\circ}, \tau_{\circ}$ such as rated velocity and torque with coefficients $\alpha_i, \beta_i$. Next, the vertices are associated with the motor library through a binary combination, $\bmm{v}_i(\bmm{x}) = (\alpha_i\omega_\circ(\bmm{x})_B, \beta_i\tau_\circ(\bmm{x})_B)$. Consequently, the operation domain is expressed as follows,
\begin{equation}
    P(\bmm{x}, \bmm{\alpha}, \bmm{\beta}) = \convhull \big(\bmm{v}_1(\bmm{x})_B, \dots, \bmm{v}_k(\bmm{x})_B \big),
\end{equation} 
where $\bmm \alpha$ and $\bmm \beta$ are vectors of coefficients $\alpha_{(\cdot)}$ and $\beta_{(\cdot)}$. 

Next, the joint-space trajectories of given tasks are concatenated and projected to motor-space. The joint-space trajectory of length $T$, $U^{\tj}=\{(\bmm{\omega},\bmm{\tau})_k\:|\: \bmm{\omega}, \bmm{\tau} \in \mathbb{R}^m, k\in[T]\}$, is projected to actuator-space using actuator-joint coupling Jacobian $\bmm{C}(\bmm{z})_B$, then projected to motor-space using gear ratio Jacobian $\bmm{N}(\bmm y) = \diag(N_1(\bmm y)_B,\dots,N_m(\bmm y)_B)$. The tasks trajectory on motor-space $U^{\textrm{m}}(\bmm y,\bmm z)$ is obtained 
as a function of choices of gearboxes and couplings,
\begin{equation}
\begin{aligned}
    U^{\tm}(\bmm y,\bmm z) = \{(&\bmm{\omega}, \bmm{\tau})  \:|\: \bmm{\omega} = \bmm{N}\bmm{C}^{-1}\bmm{\omega}_\tj, \\  &\bmm{\tau}=\bmm{N}^{-\top}\bmm{C}^\top \bmm{\tau}_\tj, 
     (\bmm{\omega}_\tj, \bmm{\tau}_\tj) \in U^\tj\}.
\end{aligned}
\end{equation}
Finally, the following task feasibility constraint ensures the task trajectory to be contained within the operation domain 
\begin{equation}
    U^m(\bmm{y},\bmm{z}) \subseteq P(\bmm{x}, \bmm{\alpha}, \bmm{\beta}),
\end{equation}
which easily converts to a set of linear constraints. 

\subsection{Design Rules} 
In this study, we bound the motor mass budget. If the motor mass budget is left unbounded, the optimizer will always prefer heavier, larger, and more torque-dense motors. However, there is a limit to the diameter of the motor that fits inside the robot's body. Moreover, the sum of motor mass may take up to $40\%$ of the entire mass of a robot \cite{minitaur}. Hence, a smaller motor mass budget may leave significant room for a larger payload. The motor mass budget constraint is enforced as follows,
\begin{equation}
    m_t(\bmm{x})_B = \tsum_{i=1}^m m_i x_i \le m_t^{\textrm{UB}},
\end{equation}
where $m_t$ is total mass of a design case, $m_i$ is mass of candidate motors and $m_t^{\textrm{UB}}$ is the total mass budget.

Designers can further employ various design rules such as limits on torque, power, etc. at various levels such as motor, actuator, and joint-levels. It is because system capability and task trajectories can be explicitly expressed using binary combinations. Second, logical constraints are also allowed. A statement, 'If a component $A_1$ is used, then $A_2$ also is used', is translated as $x_1 - x_2 \ge 0$ where $x_1$, $x_2$ are binary variables that associate components $A_1$, $A_2$. For more logical constraints, refer to \cite{schrijver1998theory}.

\section{Humanoid Design Study}

\subsection{Robot Model and Tasks}
A humanoid robot is modeled as a single rigid body with massless limbs and weight $20$\,kg that closely resembles Tello in \cite{sim2022tello}. The joint ordering is hip yaw, hip roll, hip pitch, and knee pitch followed by ankle pitch. 

\begin{figure}[t]
     \centering
     \subfloat[]{
        \label{fig:walk}
        \includegraphics[width = 0.7\linewidth]{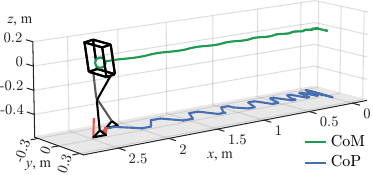}}
        \\
     \subfloat[]{
        \label{fig:snatchTask}        \includegraphics[width = 0.75\linewidth]{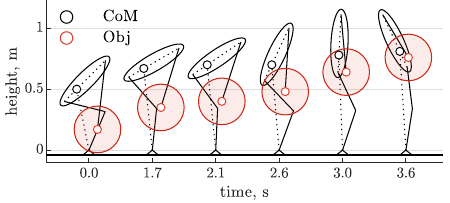}}
    \caption{Sketches of tasks used in humanoid design study. (a) walking forward at moderately fast speed and (b) lifting an object in snatch style. }
    \label{fig:tasks}
\end{figure}

Next, based on the ROM, two task trajectories were generated (Fig.\,\ref{fig:tasks}). The first task corresponds to moderately fast walking, with a peak speed of 1.3\,m/s, a step frequency of 0.3\,s. Joint torques and velocities of single support phase (0.15\,s) are used. The walking trajectories are generated by simulating the SRBD of Tello. A step placement feedback controller that tracks a predefined walking speed is implemented \cite{gong2021one}. An optimization-based balance controller, similar to in \cite{chignoli2020variational}, computes the ground reaction forces required to realize the desired center-of-mass walking dynamics. The SRBD task-space trajectories are converted to joint-space trajectories using inverse kinematics. The resulting joint velocities, which contain high-frequency signals, are low-pass filtered at a cutoff frequency of 30\,Hz to obtain meaningful solutions. The second task trajectory is to lift a point mass of 15\,kg in a snatch style 
over 3.6\,s. This trajectory is generated by extracting the joint trajectory from a video of an athlete performing a snatch. The trajectory is scaled down to match the size of the robot, and joint torques are obtained using inverse dynamics. The lifting task only recruits the pitch axes of the hip, knee, and ankle joints.

It is important to note that designers may choose other tasks and this study is merely showcasing an example. Depending on the choice of task, optimal designs may vary significantly.

\subsection{Design Library and Rules}
The design space is formulated by combinations of the following component libraries. The motor library contains $n_m=10$ motors from CubeMars (AK,\,R,\,RI series \cite{cubeMars}). The reduction library consists of ideal transmission without frictional loss and gear ratio ranges from 1 to 12 at whole number interval $(n_r = 12)$. The vertices of the operation domains of the motors are defined by peak torque and rated velocity from datasheets and vectors of tuning coefficients $\bmm{\alpha} =(1.5, -2.5, -2.5, 1.5), \, \bmm{\beta} = (1.2, 1.2, -1.2, -1.2)$. 

\begin{table}[t]
\begin{center}
\footnotesize
\renewcommand{\arraystretch}{1.1}
\begin{threeparttable}
\caption{List of Actuator-Joint Couplings}
\begin{tabular}{rrrrr}
\toprule
Name & Jacobian & & Name & Jacobian
\\
\cmidrule(lr){1-2} \cmidrule(lr){4-5}
serial & $\diag({1, 1, 1, 1, 1})$ & & par-45 & $\diag(1, 1, 1, \bmm{C}_d)$
\\
par-12 & $\diag(\bmm{C}_d, 1, 1, 1)$ && par-12-34 & $\diag(\bmm{C}_d, \bmm{C}_d, 1)$
\\
par-23 & $\diag(1, \bmm{C}_d, 1, 1)$ && par-12-45 & $\diag(\bmm{C}_d, 1, \bmm{C}_d)$
\\
par-34 & $\diag(1, 1, \bmm{C}_d, 1)$ && par-23-45 & $\diag(1, \bmm{C}_d, \bmm{C}_d)$
\\
\bottomrule 
\label{table:Jacobians}
\end{tabular}
\end{threeparttable}
\end{center}
\end{table}

There are 8 couplings studied in this paper including 1 serial and 7 parallel couplings. The parallel couplings are based on modules of 2-DoF differential coupling whose Jacobian is
\begin{equation}
    \bmm{C}_d = \frac{1}{2}\begin{bmatrix}
        1 & 1 \\ -1 & 1
    \end{bmatrix}.    
\end{equation}
The couplings are labeled as `serial', 'par-($n_1m_1$)' and `par-($n_1 m_1$)-($n_2 m_2$)' where $n_i, m_i$ denotes the index of joints involved in $i$'th differential coupling (see Table.\,\ref{table:Jacobians}). Since a mechanical coupling between faraway joints is typically impractical, the couplings are limited to consecutive joints. The hardware realization of Tello \cite{sim2022tello} utilizes `par-23-45'.

The total mass of 5 motors is subject to a constraint (motor mass budget), with 25 values ranging from 2.2\,kg to 4.0\,kg at intervals of 0.075\,kg. After a few runs of modifying the values of the motor mass budget, it was found that the optimization was infeasible below 2.2\,kg and solutions did not vary beyond 4.0\,kg.

\subsection{Design Optimization}
The optimization problem is detailed as follows. Inclusion of library data is implied by the use of binary optimization variables $\bmm x,\bmm y,\bmm z$. 
\begin{mini}|s|
{\bmm{x}, \bmm{y}, \bmm{z} }{\tsum_{k=1}^{T}\trace(\overline{\bmm{H}}^{\tj}_r (\bmm{x},\bmm y, \bmm z, k))}   
{\label{eqn:optiDetail} }{}
\addConstraint{
U^m(\bmm{y},\bmm{z}) \subseteq P(\bmm{x}, \bmm{\alpha}, \bmm{\beta})
}
\addConstraint{
m_t(\bmm{x})_B  \le m_t^{\textrm{UB}}
}
\addConstraint{
\bmm{x}\in\mathbb{B}^{n_a \times n_m}, \bmm{y}\in\mathbb{B}^{n_a \times n_r}, 
\bmm{z}\in\mathbb{B}^{n_c}
}
\addConstraint{
\textrm{exclusivity}(\bmm x,\bmm y,\bmm z)
}
\end{mini} 
To explicitly study the effect of constraining total mass budget for all 5 motors and employing different coupling Jacobian, the optimization was run for all combination of mass budget constraints and coupling Jacobians (total $25 \times 8 = 200$ runs). 
After optimal solutions are obtained, system characteristics such as inertia matrix and force/torque capability polytopes (FCP/TCP) in joint and task-space and copper loss on actuators are retrieved. The FCP/TCP represents the
maximum force/torque that can be applied for a given configuration, based on the limits of actuators  \cite{chiacchio1997force}. 

The optimization problem \eqref{eqn:optiDetail} is linearized using techniques introduced in Sec.\,\ref{sec:background}. The design optimization is written in MATLAB and calls an integer linear programming solver (\texttt{intlinprog}, Gurobi). The solver was simultaneously run on CPU (20 cores, Intel i7-12700H) with default solver parameters. Computation of total 200 runs took less than 100\,s.

\section{Results and Discussion}

\begin{figure}[t]
     \centering
     \subfloat[]{
        \label{fig:mass-trace-loss} \includegraphics[width=0.98\linewidth]{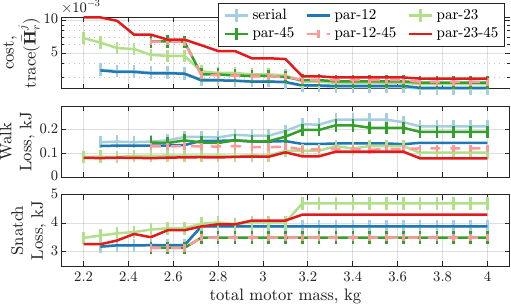}}
        \\
    \subfloat[]{
        \label{fig:pareto3D}        \includegraphics[width=0.85\linewidth, clip, trim=0mm 4mm 0mm 0mm]{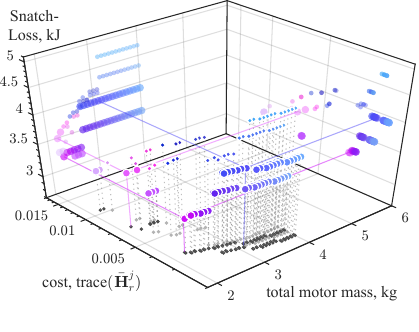}}
    \caption{(a) Optimal costs (trace of joint-space reflected inertia) vs. total mass budget for 5 motors for every coupling case. Total copper losses per walking and snatch lifting task (b) Optimal designs' motor mass budget, optimal cost and copper loss of snatch task. The bigger dots represent Pareto front. }
    \label{fig:result1}
\end{figure}

\begin{figure*}[t]
    \centering
    \includegraphics[width=0.99\linewidth, clip, trim=0mm 133mm 177mm 0mm]{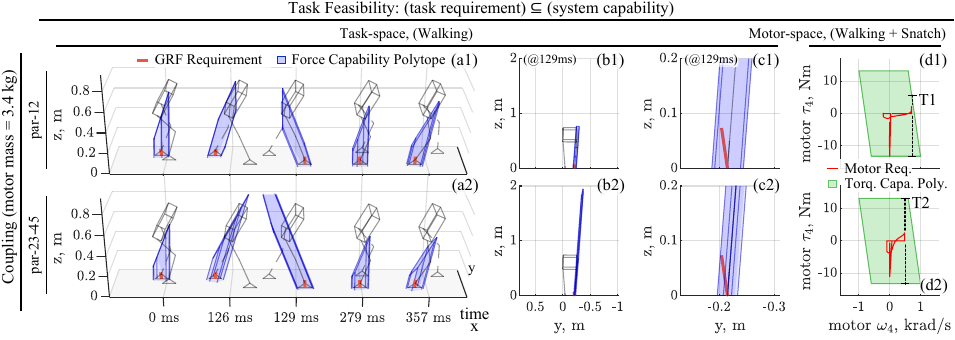}
    \caption{Comparison of two optimal designs under identical motor mass budget and with different usage of coupling mechanism. (a1),\,(a2) FCP of contact leg in contact along walking. (b1),\,(b2) Frontal views show significant difference in size the FCP. (c1),\,(c2) FCPs of both designs are large enough to contain ground reaction forces to achieve moderately fast walking. (d1),\,(d2) Couplings affect how task trajectories are projected onto motor-level and motor torque capability margins (T1 vs. T2). }
    \label{fig:taskFeasibility}
\end{figure*}

\subsection{Motor Mass Budget vs. Reflected Inertia vs. Copper Loss}
The first row of Fig.\,\ref{fig:result1}(a) gives an overview of optimal designs obtained with varied motor mass budgets and use of different couplings. First, as the motor mass budget decreases, availability of larger and more torque-dense motors decreases. Consequently, the optimal solutions leans towards smaller and weaker motors, necessitating the usage of elevated gear ratios. In the end, the use of higher gear ratios inflates reflected inertia or contributes to higher cost (optimal objective function value). 

As to the effect of couplings, a few coupling cases were found infeasible (par-34, par-12-34) across all motor mass budgets, whereas other couplings become infeasible as the motor mass budget decreases. When the motor mass budget was limited to extremely small values, only coupled mechanisms provided sufficient torques and velocities (feasible). The serial actuation (no coupling) demonstrated the least reflected inertia at the cost of high copper loss for walking task across all cases of motor mass budget. In conclusion, this result suggests that 1) the framework can filter out couplings that cannot generate sufficient torque or speed, 2) at extremely low motor mass budget, actuators need to cooperate so that they can amplify joint torques to achieve desired tasks, and 3) use of coupling increases the reflected inertia for designs with a large motor mass budget. 

The second and third row of Fig.\,\ref{fig:result1}(a) shows the combined copper loss of all motors of each task. As the motor mass budget decreases, the copper loss also decreases for most couplings. This tendency is opposite to that of reflected inertia because larger reflected inertia implies higher gear ratios, which again implies less motor torque and motor current draw. Two optimal designs following this trend are compared in Table.\,\ref{table:DesignComparison}. However, these implications are not strict: in the walking task, a few coupling cases did not follow this tendency (Fig.\,\ref{fig:mass-trace-loss}, second row, serial, par-45, par-23-45, around 3.2$\sim$3.7 kg). This is because different choices of gear ratio and motor can emerge as long as they are optimal and satisfy design rules.

If this study is posed as a multi-objective optimization with minimization of reflected inertia and the copper loss, the optimal solutions can be plotted in 3D space as Fig\,\ref{fig:result1}(b). Pareto optimal solutions are emphasized as larger dots over other optimal solutions. To aid visual understanding, optimal solutions are projected onto 2D slices.  

\subsection{Effect of Actuator-Joint Coupling}

The influence of the usage of couplings on task-space FCP and motor-level trajectory is visualized in Fig.\,\ref{fig:taskFeasibility} which presents a comparative analysis between two design cases, specifically, `par-12' and `par-23-45'. The latter utilizes more a higher degree of coupling than the former while both are designed with an identical motor mass budget.

A notable point in this comparison involves the velocity-dependent nature of FCP, attributed to the contour of the operational domain of motors. When motor velocities are high, the available torque (T1) near the boundary of the operation domain is reduced due to high motor velocity (Fig.\,\ref{fig:taskFeasibility}(d1),\,(d2)). 

The latter case, with additional degrees of coupling, has more chances for cooperative actuation which means that the load on a joint can be shared among the coupled actuators. Thanks to the load sharing, the individual actuators experience lower torque requirements. This allows employing smaller gear ratios which renders smaller motor velocities. Hence, employing more degrees of coupling has a benefit of keeping the motor-level trajectory (velocity) at a safe distance from the boundary of the operational domain. 
The degradation of motor torque capability in design case `par-12' 
manifests as a significantly smaller FCP at 129\,ms (Fig.\,\ref{fig:taskFeasibility}(a1),\,(b1)), in contrast to a larger FCP in (Fig.\,\ref{fig:taskFeasibility}(a2),\,(b2)).

The state-dependent nature of the FCP could be related to the margin of admissible error of feedback controllers. For instance, if joint velocity deviates from a (pre)planned trajectory towards the boundary of the operation domain of motors, the torque capability diminishes. Simultaneously, a feedback controller will attempt to bring the state back to reference. At that moment, if the torque margin is insufficient, the torque commands from the feedback controller could suffer from saturation which may lead to failure. Designers can prevent this by modifying the approximated operation domain to be smaller than the actual operation domain of motors by tuning the coefficients $\bmm{\alpha}$ and $\bmm{\beta}$. 

\begin{table}[t]
\begin{center}
\footnotesize
\renewcommand{\arraystretch}{1.2}
\begin{threeparttable}
\caption{Comparison of Two Optimal Design Cases}
\begin{tabular}{@{}r*{5}{R{24pt}}}
\toprule
 & \multicolumn{5}{c}{Joint (Axis)}
\\ \cmidrule(lr){2-6}
& Hip(z)    & Hip(x)    & Hip(y)    & Knee(y)   & Ankle(y)
\\ 
\cmidrule(lr){1-6}
\multicolumn{6}{l}{\textbf{Design Case A} \textit{(lightest, coupling}~$=$~(par-12), $m_t=2.28$~kg\textit{)}}
\\
Inertia [$\textrm{gm}^2$] & \multicolumn{2}{r}{(~0.31, 0.31~)} & 1.06 & 1.06 & 0.75
\\
Peak~Torq.~[Nm] &  \multicolumn{2}{r}{(~25.7, 25.7~)}  &  92.4  &  92.4  &  28.9
\\
Motor&  \multicolumn{2}{r}{(~RI70, RI70~)}    &   R100    &   R100    &   RI70
\\
Gear Ratio&     \multicolumn{2}{r}{(~3, 5~)}  &   7  &   7  &   9
\\
\addlinespace[1.5ex] \multicolumn{6}{l}{\textbf{Design Case B} {\textit{(heaviest, coupling}~$=$~(par-23-45), $m_t=3.7$~\textrm{kg}\textit{)}}}
\\ 
Inertia [$\textrm{gm}^2$] & 0.068 &  \multicolumn{2}{r}{(~0.69, 0.69~)} & \multicolumn{2}{r}{(~0.73, 0.73~)}
\\
Peak~Torq.~[Nm] &  26.4 & \multicolumn{2}{r}{(~105.6, 105.6~)}    &  \multicolumn{2}{r}{(~105.6, 105.6~)}    
\\
Motor&   R100    &   \multicolumn{2}{r}{(~R100, R100~)}    & \multicolumn{2}{r}{(~R100, R100~)}    
\\
Gear Ratio&     2  &    \multicolumn{2}{r}{(~4, 4~)}    &      \multicolumn{2}{r}{(~3, 5~)}
\\
\bottomrule \label{table:DesignComparison}
\end{tabular}
\vspace{-10pt}
\begin{tablenotes}[flushleft]
\item Note: Two entities within a parenthesis denotes two joints involved in 2-DoF actuator-joint coupling. 
\end{tablenotes}
\end{threeparttable}
\end{center}
\end{table}

\subsection{Characterization of Inertia and Capability Polytopes} \label{sec:task_joint}
\begin{figure}[t]
\centering
\includegraphics[width=0.99\linewidth, clip, trim=0mm 141mm 252mm 0mm]{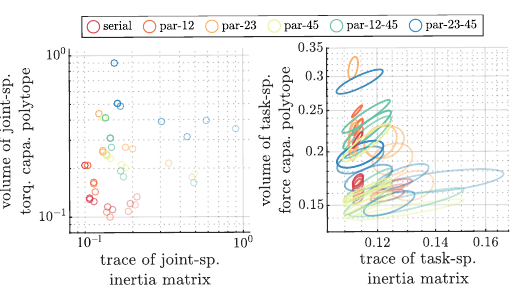}
\caption{Normalized optimal objective function value (trace of reflected inertia matrix) vs. normalized volume of FCP/TCP in joint-space (left) and task-space (right).}
\label{fig:inertia_volFCP}
\end{figure}

The relation between reflected inertia and FCP/TCP in both joint- and task-space is analyzed to address two possible claims; 1) minimization of reflected inertia should take place at task-space level, not joint-space and 2) maximization of TCP/FCP could be also taken into account as well as minimization of the reflected inertia. Unfortunately, the authors could not obtain direct quantification of reflected inertia in task-space and the volume of TCP/FCP using binary integer linear programming. Still, a few trends were observed by plotting the relation between the size of reflected inertia versus volume of TCP/FCP in joint-space and task-space as in Fig.\,\ref{fig:inertia_volFCP}. Optimal designs with smaller motor mass budgets are plotted with more opacity. The TCP/FCP were calculated without considering the degradation in torque margin due to high velocity. Since FCP varies as configuration (joint angles) changes, its distribution is illustrated as an 0.7-sigma ellipse. In each space, two similar trends were observed; 1) serial actuation shows more cohesiveness in the relation between the size of reflected inertia and FCP/TCP compared to coupled ones, 2) a combination of coupled actuation and smaller mass budget leads to larger size and variation of reflected inertia, and 3) larger motor mass budget allows larger FCP/TCP.

\section{Conclusion} 

The existing design principles and optimal design tools for the design of multi-DoF actuation systems often fall short in providing practical designs or understanding of optimal designs. To resolve this issue, an interactive optimal design tool for actuator system for humanoids is introduced. This tool is designed to incorporate expert designers' knowledge and rapidly solve large numbers of design studies to expand understanding of optimal designs.

This paper closes with several limitations of the present study and suggestions for future research. First, ROM employed in this study assumes massless limbs and actuator dynamics is neglected. Hence, the inertial forces exerted on the limbs and rotors within the actuation system are not accounted for. Future work may incorporate inertial forces associated with the limbs' apparent inertia and the reflected inertia of the actuators into the optimization process. 
Second, the applicability of optimal design solutions produced by the proposed tool remains unclear. An essential next step would be to integrate this tool with simulation software for the validation of proposed designs.




\section*{ACKNOWLEDGMENT}
To research advisors Patrick Wensing and Donghyun Kim, Johannes Englsberger, and my colleagues Donghoon Baek and Seunghyun Bang  for guidance and fruitful discussion.

\bibliographystyle{IEEEtran}
\bibliography{main.bib}

\end{document}